\documentclass{article}

\PassOptionsToPackage{numbers}{natbib}

     \usepackage[preprint]{neurips_2020}

\usepackage[utf8]{inputenc} 
\usepackage[T1]{fontenc}    
\usepackage{hyperref}       
\usepackage{url}            
\usepackage{booktabs}       
\usepackage{amsfonts}       
\usepackage{nicefrac}       
\usepackage{microtype}      

\usepackage{soul}
\usepackage{color}
\usepackage{algorithm}      
\usepackage[noend]{algpseudocode} 
\usepackage{amsmath}

\DeclareMathOperator*{\argmin}{arg\,min}
\usepackage{graphicx}
\usepackage{subcaption}

\usepackage[export]{adjustbox}
\usepackage{lineno}
\usepackage{enumitem}

\usepackage{setspace}

\usepackage{todonotes}
\usepackage{lipsum}

\title{Complex Robotic Manipulation via Graph-Based Hindsight Goal Generation}

\author{%
	Zhenshan Bing{$^*$}\\
	Department of Informatics\\
	Technical University of Munich, Germany\\
	\texttt{bing@in.tum.de} \\
	\And
	Matthias Brucker{$^*$} \\
	Munich School of Engineering\\
	Technical University of Munich, Germany\\
	\texttt{matthias.brucker@tum.de} \\
	\And
	Kai Huang \\
	School of Data and Computer Science\\
	Sun Yat-sen University, China\\
	\texttt{huangk36@mail.sysu.edu.cn} \\
	\And
	Fabrice O. Morin, Alois Knoll \\
	Department of Informatics\\
	Technical University of Munich, Germany\\
	\texttt{\{morinf, knoll\}@in.tum.de} \\
	\And
}

\begin{document}

\maketitle

\begin{abstract}
	Reinforcement learning algorithms such as hindsight experience replay (HER) and hindsight goal generation (HGG) have been able to solve challenging robotic manipulation tasks in multi-goal settings with sparse rewards. 
	HER achieves its training success through hindsight replays of past experience with heuristic goals, but under-performs in challenging tasks in which goals are difficult to explore.
	HGG enhances HER by selecting intermediate goals that are easy to achieve in the short term and promising to lead to target goals in the long term.
	This guided exploration makes HGG applicable to tasks in which target goals are far away from the object's initial position. 
	However, HGG is not applicable to manipulation tasks with obstacles because the euclidean metric used for HGG is not an accurate distance metric in such environments.
	In this paper, we propose graph-based hindsight goal generation (G-HGG), an extension of HGG selecting hindsight goals based on shortest distances in an obstacle-avoiding graph, which is a discrete representation of the environment. 
	We evaluated G-HGG on four challenging manipulation tasks with obstacles, where significant enhancements in both sample efficiency and overall success rate are shown over HGG and HER. 
	Videos can be viewed at 
	\href{https://sites.google.com/view/demos-g-hgg/}{https://sites.google.com/view/demos-g-hgg/}.
\end{abstract}

%
\section{Introduction}

In recent years, deep reinforcement learning (RL) has made significant progress, with RL concepts being successfully applied to decision-making problems in robotics, which include, but not limited to, helicopter control \cite{Ng2006}, hitting a baseball \cite{Peters2008}, door opening \cite{Chebotar2017}, screwing a cap onto a bottle \cite{Levine2016}, and object manipulation \cite{Andrychowicz2017}. 
To train a meaningful policy for such tasks, neural networks are used as function approximators for learning a value function to optimize a long-term expected return. Estimation of the expected return is based on a reward function, which \textcolor{black}{is highly associated with the tasks and therefore }must be thoroughly shaped for policy optimization.

In most real-world applications, where a concrete representation of efficient or even admissible behavior is unknown, the design of a reward function is challenging and time-consuming, thereby hindering the wide applications of RL.
Consequently, algorithms \textcolor{black}{that can} support learning from sparse rewards are \textcolor{black}{highly desirable}, e.g., a binary signal indicating successful task completion; since sparse rewards are easy to derive from the task definition without further engineering.
Andrychowicz et al.~\cite{Andrychowicz2017} introduced an algorithm called `` hindsight experience replay (HER)'', which improves the success of off-policy RL algorithms in multi-goal RL problems with sparse rewards. 
The concept behind HER is to first learn with \textcolor{black}{hand-crafted heuristic} intermediate goals that are easy to achieve, and then continue with more difficult goals. 
Precisely, HER constructs hindsight goals from previously achieved states, replays known trajectories with these hindsight goals, and trains the goal-dependent value function based on the results. 
While HER has proven to work efficiently in environments where goals \textcolor{black}{can be easily reached through random explorations}
\cite{Plappert2017}, it fails to reach goals that are far away from initial states and \textcolor{black}{hard to reach}. 
Due to random exploration and the heuristic choice of hindsight goals from achieved states, hindsight goals keep being distributed around the initial state, far away from the target goals, which will never be reached since no reward signal is obtained. 

Hindsight goal generation (HGG)~\cite{Ren2019} tackles the aforementioned problem by using intermediate hindsight goals as an implicit curriculum to guide exploration towards target goals. 
HGG aims at choosing hindsight goals that are both easy to achieve and challenging enough to help the function approximator learn how to achieve the target goals eventually. 
In HGG, the choice of hindsight goals is based on two criteria: the current value function (as much knowledge as possible about how to reach the hindsight goals) and the Wasserstein distance between the target goal distribution and the distribution of potential hindsight goals (goals as close as possible to the target goal distribution). 
The resulting Wasserstein-Barycenter problem is discretely solved using the euclidean distance metric between two goals sampled from the potential hindsight goal distribution and the target goal distribution.
While HGG demonstrates higher sample efficiency than HER in environments where the euclidean metric is applicable, it fails in environments with obstacles, where the shortest obstacle-avoiding distance between two goals cannot be computed with the euclidean metric.

In this paper, we introduce graph-based hindsight goal generation (G-HGG), an extension of HGG, designed for sparse-reward RL in robotic object manipulation environments with obstacles. 
To make G-HGG applicable to environments with obstacles, the euclidean metric is replaced by a graph-based distance metric: the environment's goal space is represented by a graph consisting of discrete goals as vertices and obstacle-avoiding connections between the vertices as edges. 
The distance between two goals are then approximated by the shortest path on the graph between the two vertices that are closest to the two goals.
For verification, we evaluated G-HGG on four different challenging object manipulation environments. 
With the graph-based distance metric, a significant enhancement in both sample efficiency and overall success rate is demonstrated over HGG and HER. Ablation studies in the appendix (See supplementary material) show the robustness of G-HGG in terms of discretization density.
\section{Background}

\textbf{Goal-Conditioned RL~} 
\textcolor{black}{
In goal-conditioned RL, an agent interacts with its environment to reach some goals, which can be modeled as a goal-conditioned Markov decision process (MDP) with a state space $\mathcal{S}$, an action space $\mathcal{A}$, a goal space $\mathcal{G}$, a probabilistic transition function $P : S \times \mathcal{A} \rightarrow \mathcal{S}$, a reward function $r_g: \mathcal{S} \times \mathcal{A} \rightarrow \mathbb{R}$, and a discount factor $\gamma$.  
At every time step $t$, the agent's behavior $a_t$ is defined by a 
probabilistic
policy $\pi(s_t||g)$, given by the current state $s_t$ and the goal $g$
(we use $||$ as a symbol for concatenation into $\mathcal{S} \times \mathcal{G}$).
The task is to find an optimal policy that can maximize the expected curriculum reward starting from the initial state sampled from the initial state distribution $s \in S_0$, which is defined as 
}
\begin{equation}
V^{\pi}(s||g) = \mathbb{E}_{s_0 = s \sim S_0,\, a_t \sim \pi(s_t||g),\, s_{t+1} \sim P(s_t, a_t)}\big[\sum_{t=0}^{\infty} \gamma^t r_g(s_t, a_t)\big].
\label{eq:V}
\end{equation}

\textbf{Hindsight goal generation}
HGG \cite{Ren2019} is an extension of HER~\cite{Andrychowicz2017} to scenarios in which the target goal distribution differs a lot from the initial state distribution.
HGG can be briefly explained as follows.
%
A value function of a policy $\pi$ for a specific goal $g$ is assumed to have some generalizability to another goal $g'$ close to $g$~\cite{Asadi2018,Luo2019}. 
This assumption is mathematically characterized via Lipschitz continuity as
\begin{equation}
\label{eq:Lipschitz}
|V^\pi(s||g) - V^\pi(s'||g')|\leq L \cdot d(s||g,s'||g'), 
\end{equation}
where $d(s||g,s'||g')$ is a metric defined by
\begin{equation}
d((s||g),(s'||g')) = c||m(s)-m(s')||_{\textcolor{black}{2}} + ||g-g'||_{\textcolor{black}{2}}.
\end{equation}
$m(\cdot)$ is a state abstraction to map from the state space to the goal space.
\textcolor{black}{The hyper-parameter $c > 0$ provides a trade-off between 1) the distance between target goals and 2) the distance between the goal representation of the initial states.}
\textcolor{black}{Assuming} the generalizability condition \eqref{eq:Lipschitz} holds for two distributions $\textcolor{black}{s^{(1)}_0}||\textcolor{black}{g^{(1)}} \sim \mathcal{T}^{\textcolor{black}{(1)}}$ and $\textcolor{black}{s^{(2)}_0}||g^{\textcolor{black}{(2)}} \sim \mathcal{T}^{\textcolor{black}{(2)}}$, Ren et al.~\cite{Ren2019} demonstrated that
	\begin{equation}
	\label{eq:generalizability}
	V^\pi(\mathcal{T}^{\textcolor{black}{(2)}}) \geq V^\pi(\mathcal{T}^{\textcolor{black}{(1)}}) - L \cdot \mathcal{D}(\mathcal{T}^{\textcolor{black}{(1)}}, \mathcal{T}^{\textcolor{black}{(2)}}),
	\end{equation}
%
%
	where $\mathcal{D}(\cdot,\cdot)$ is the Wasserstein distance~\cite{ruschendorf1985wasserstein} based on $d(\cdot, \cdot)$, defined as
	\begin{equation}
	\mathcal{D}(\mathcal{T}^{(1)}, \mathcal{T}^{(2)}):= \inf_{\mu \in \Gamma(\mathcal{T}^{(1)}, \mathcal{T}^{(2)})}\bigg(\mathbb{E}_\mu \Big [d(s_0^{(1)}||g^{(1)}, s_0^{(2)}||g^{(2)})\Big]\bigg ).
	\end{equation}
	$\Gamma(\mathcal{T}^{(1)}, \mathcal{T}^{(2)})$ denotes the collection of all joint distributions $\mu(s_0^{(1)} || g^{(1)}, s_0^{(2)}|| g^{(2)})$, \textcolor{black}{the marginal probabilities of which} are $\mathcal{T}^{(1)}$ and $\mathcal{T}^{(2)}$.
\textcolor{black}{With $\mathcal{T}^*$ denoting the joint distribution over initial state $s_0$ and goal $g$,	
the agent} tries to find a policy $\pi$ maximizing the expectation of the discounted cumulative reward based on the state value function defined in \eqref{eq:V}:
\begin{equation}
	\label{eq:maxreward}
	V^\pi(\mathcal{T}^*):= \mathbb{E}_{s_0||g \sim \mathcal{T}^*}\Big[V^\pi(s_0||g)\Big].
\end{equation}	
From \eqref{eq:generalizability}, it can be derived that optimizing this expected cumulative reward defined in \eqref{eq:maxreward} can be relaxed into the \textcolor{black}{surrogate problem of finding}
\begin{equation}
\label{eq:optimization}
\max_{\mathcal{T},\pi}V^\pi(\mathcal{T})- L \cdot \mathcal{D}(\mathcal{T}, \mathcal{T}^*).
\end{equation}
Joint optimization of $\pi$ and $\mathcal{T}$ \eqref{eq:optimization} is solved in a two-stage iterative algorithm.
First, standard policy optimization maximizes the value function $V^{\pi}$ based on experience generated from the intermediate task set $\mathcal{T}$. 
Second, the intermediate task set $\mathcal{T}$ is optimized while the policy $\pi$ is kept constant. 
The second step is a variant of the Wasserstein Barycenter problem with the value function as a bias term for each initial state-goal pair, which can be solved as a bipartite matching problem \cite{Duan2012}. 
For this to work, $\mathcal{T}^*$ is approximated by by $K$ initial state-goal pairs sampled from it, resulting in $\hat{\mathcal{T}}^* = \{(\hat{s}_0^i, \hat{g}^i)\}_{i=1}^K$. 
Now, for every $(\hat{s}_0^i, \hat{g}^i) \in \hat{\mathcal{T}}^*$, a trajectory $\tau^i=\{s_t^i\}$ from the replay buffer is found that minimizes 
\begin{equation}
\label{eq:wasserstein}
w((\hat{s}_0^i, \hat{g}^i), \tau^i):= c ||m(\hat{s}_0^i)-m(s_0^i)||_{\textcolor{black}{2}} + \min_t \big(||\hat{g}^i - m(s_t^i)||_{\textcolor{black}{2}} - \frac{1}{L}V^\pi(s_0^i|| m(s_t^i))\big).
\end{equation}
The Lipschitz constant $L$ is treated as a hyper-parameter. 
These $K$ trajectories $\tau^i$ minimize the sum 
\begin{equation}
\label{eq:wassersteinsum}
\sum_{(\hat{s}_0^i, \hat{g}^i) \in \hat{\mathcal{T}^*}} w((\hat{s}_0^i, \hat{g}^i), \tau^i)
\end{equation}	
Finally, from each of the $K$ selected trajectories $\tau^i$, the hindsight goal $g^i$ is selected from the state $s_t^i \in \tau^i$, that minimized \eqref{eq:wasserstein}. More formally,
\begin{equation}
\label{eq:hindsightgoal}
g^i = m\bigg ( \argmin_{s_t^i \in \tau_i}\Big (||\hat{g}^i - m(s_t^i)||_{\textcolor{black}{2}} - \frac{1}{L}V^\pi(s_0^i||m(s_t^i))\Big)\bigg).
\end{equation}
\textcolor{black}{
Combined with the idea of HER~\cite{Andrychowicz2017}
and replacing $(s_0,g)$ with $(\hat{s}^i_0, g^i$), the generated hindsight transition $(s_t || g, a_t, r_t, s_{t+1}|| g)$ can be then stored in the replay buffer for training the policy.
To this end, HGG is able to generate a curriculum of meaningful hindsight goals rather than hand-crafted heuristic goals from HER, guiding exploration towards target goals. 
}




\section{Methodology}
\label{chapter:methodology}



\textbf{Problem Statement:}
In this paper, we focus on solving robotic object manipulation tasks with sparse rewards, where the goal is to move an object to a certain point in 3D space with a robotic gripper arm. 
%
%
Even though the general principle of HGG is applicable, there is one major limitation. 
Solving the Wasserstein Barycenter problem in HGG consists of the computation of the distance between a goal $g \in \mathcal{G}$ and the goal representation $m(s)$ of state $s \in \mathcal{S}$. 
However, in 3D space with obstacles, the euclidean metric used in HGG is generally not applicable.
\begin{figure}[h!]
	\centering
	\begin{subfigure}[t]{.24\textwidth}
		\captionsetup{width=0.95\textwidth}
		\centering
		\includegraphics[width=1\linewidth, trim={3cm 1cm 2cm 2cm},clip]{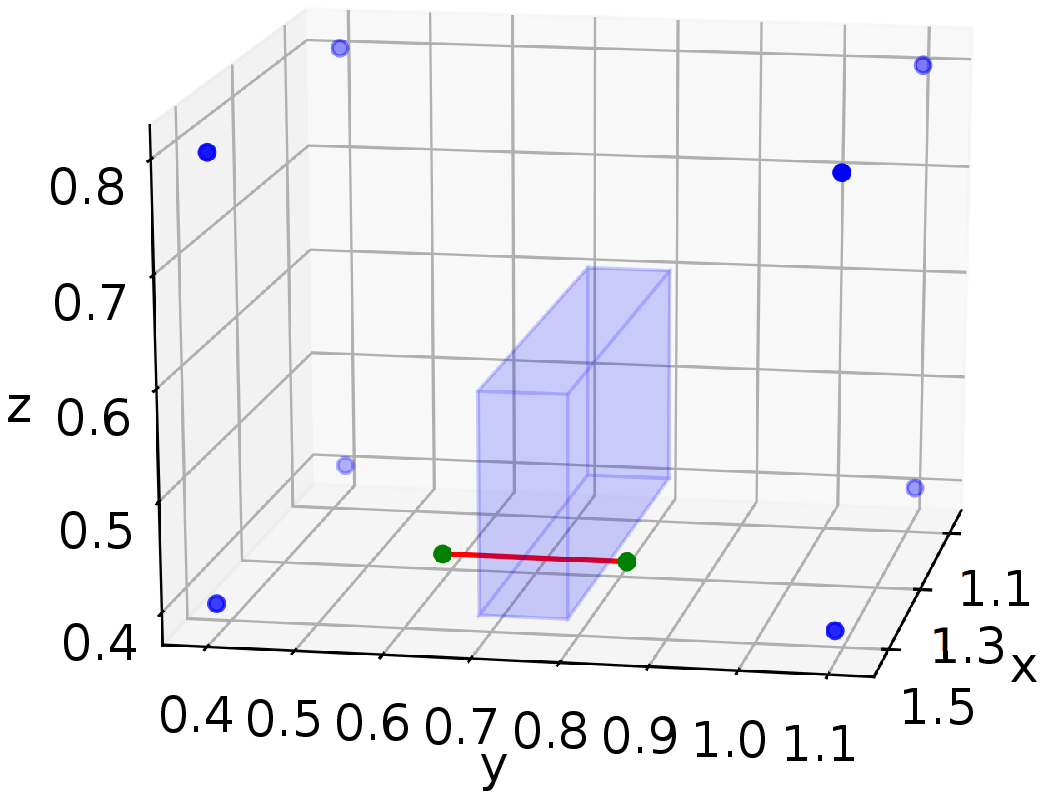}
		\caption{Case A with euclidean norm distance: Shortest path goes through the obstacle, $d_{L2}^A=0.210$}
		\label{subfig:distance_a}
	\end{subfigure}
	\begin{subfigure}[t]{.24\textwidth}
		\captionsetup{width=0.95\textwidth}
		\centering
		\includegraphics[width=1\linewidth, trim={3cm 1cm 2cm 2cm},clip]{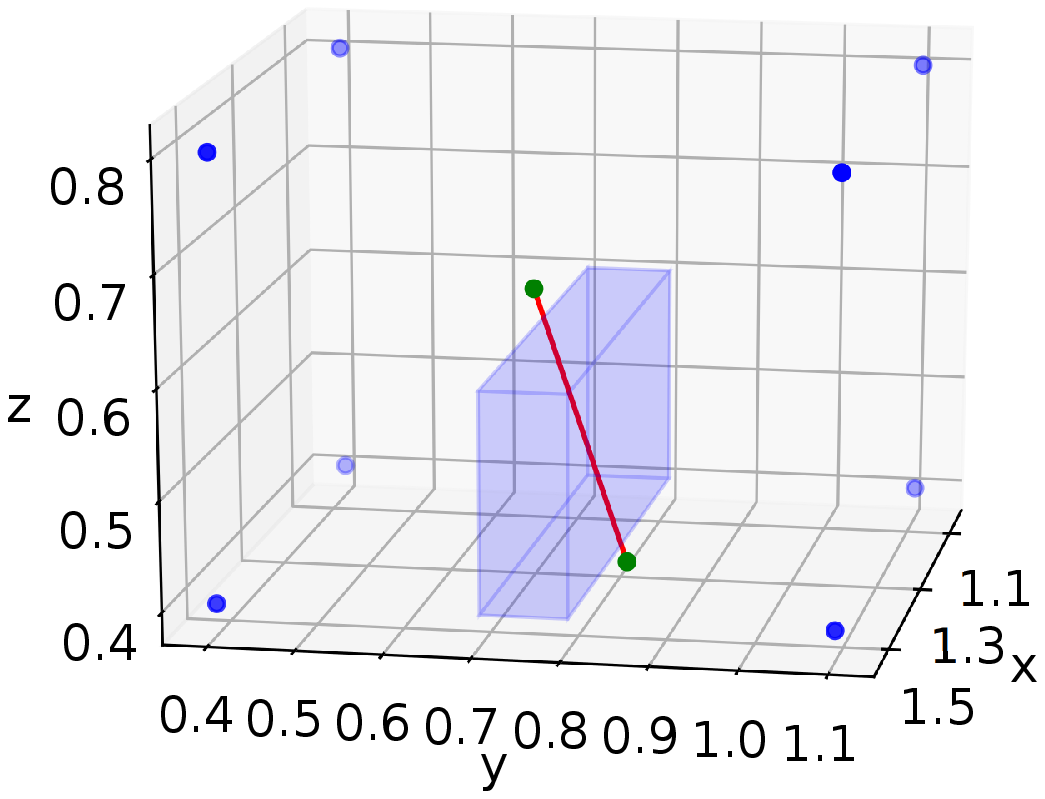}
		\caption{Case B with euclidean norm distance: Shortest path goes though the obstacle, $d_{L2}^B=0.283$}
		\label{subfig:distance_b}
	\end{subfigure}
	\begin{subfigure}[t]{.24\textwidth}
		\captionsetup{width=0.95\textwidth}
		\centering
		\includegraphics[width=1\linewidth, trim={3cm 1cm 2cm 2cm},clip]{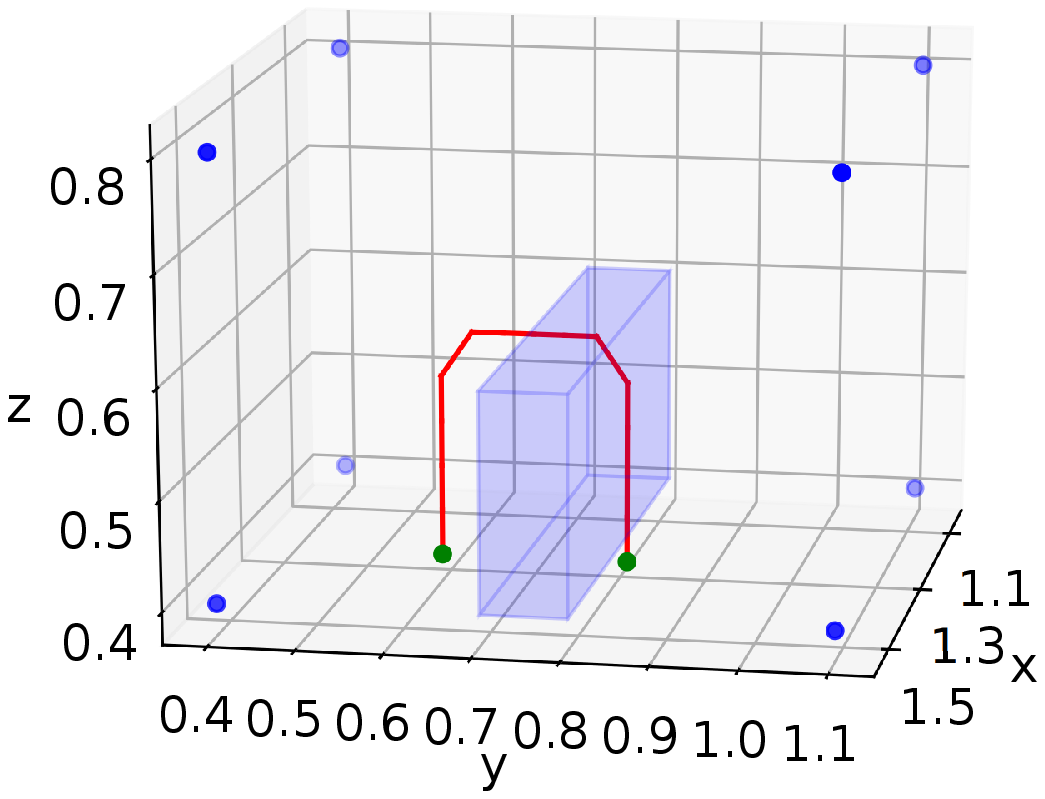}
		\caption{Case A with alternative distance: Shortest path goes around the obstacle, $d_{alt}^A=0.566$}
		\label{subfig:distance_c}
	\end{subfigure}
	\begin{subfigure}[t]{.24\textwidth}
		\captionsetup{width=0.95\textwidth}
		\centering
		\includegraphics[width=1\linewidth, trim={3cm 1cm 2cm 2cm},clip]{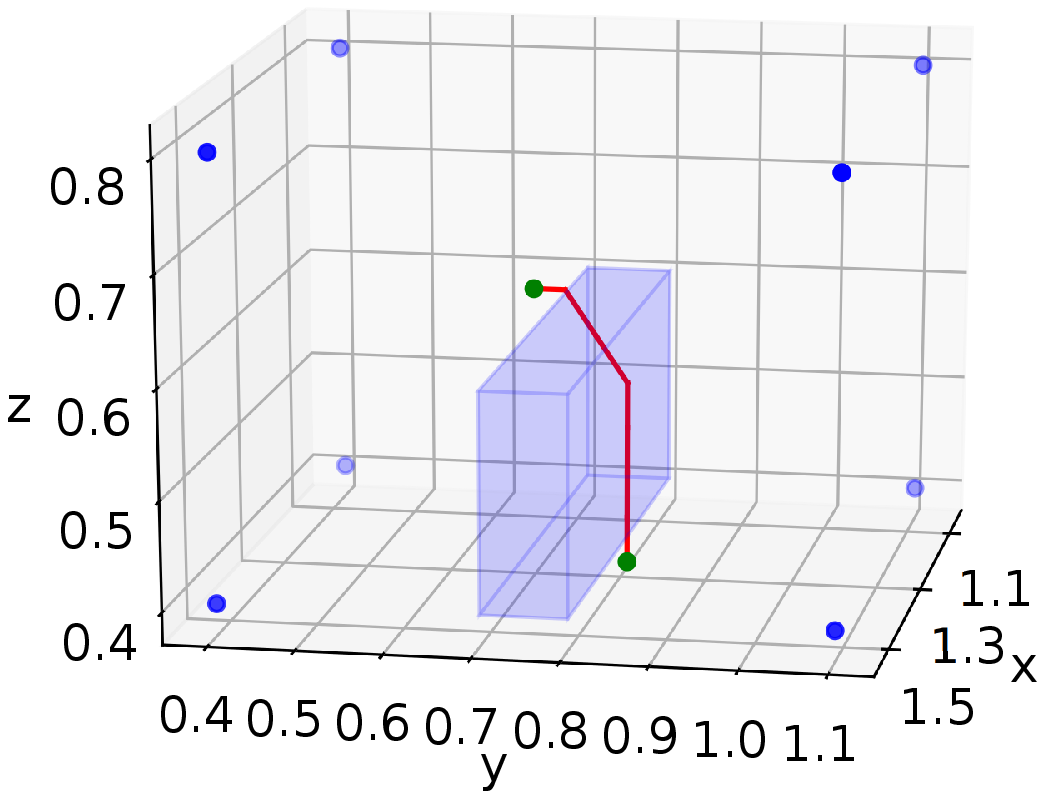}
		\caption{Case B with alternative distance: Shortest path goes around the obstacle, $d_{alt}^B=0.301$}
		\label{subfig:distance_d}
	\end{subfigure}
	\caption{Shortest distances between two points (green dots) in 3D space with obstacles.}
	\label{fig:distances}
\end{figure}
\textcolor{black}{To illustrate this point, we consider an example}, where the goal is to pick up an object \textcolor{black}{with a robotic arm}, lift it over an obstacle (blue box), and place it at the target goal. 
Let us consider two cases: A (Figures \ref{subfig:distance_a} and \ref{subfig:distance_c}) and B (Figures \ref{subfig:distance_b} and \ref{subfig:distance_d}), where goals $g_1$ and $g_2$ are marked in green. 
In Figures \ref{subfig:distance_a} and \ref{subfig:distance_b}, the euclidean metric calculates the distance, $d_{euc} = ||g_2 - g_1||$, which is marked with red lines. 
If the task is completed in an environment without obstacles, this is an accurate distance metric, and the task can be solved using HGG. 
In environments with obstacles, however, the object cannot go through these obstacles, which means the euclidean metric is not a suitable distance metric. 
Then, a suitable shortest-path based distance measure similar to the ones in Figures \ref{subfig:distance_c} and \ref{subfig:distance_d} is required in such a scenario.

 
%
%
The problem of finding the shortest path between two points without intersecting an obstacle is known as the euclidean shortest path problem.
In this paper, we propose an algorithm called graph-based hindsight goal generation (G-HGG), in which approximate shortest distances from discrete points in 3D space are pre-computed and used to compare distances between potential hindsight goals.

\subsection{G-HGG}
G-HGG is an extension of HGG to environments with obstacles, where the euclidean metric is not applicable as a distance metric. 
Hence, we reformulate \eqref{eq:wasserstein} and \eqref{eq:hindsightgoal}, replacing the euclidean metric with \textcolor{black}{the} graph-based distance metric $d_G$ .

With the new formulation, for every $(\hat{s}_0^i, \hat{g}^i) \in \hat{\mathcal{T}}^*$, we can find a trajectory from the replay buffer $\tau^i=\{s_t^i\} \in R$ that minimizes 
\begin{equation}
\label{eq:wasserstein_new}
w((\hat{s}_0^i, \hat{g}^i), \tau^i):= c ||m(\hat{s}_0^i)-m(s_0^i)||_2 + \min_t \bigg(d_G(\hat{g}^i - m(s_t^i)) - \frac{1}{L}V^\pi(s_0^i|| m(s_t^i))\bigg) \text{,}
\end{equation}
\textcolor{black}{where all these variables have the same meanings as} in \eqref{eq:wasserstein}. 
Altogether, these $K$ trajectories $\tau^i$ minimize the sum 
\begin{equation}
\label{eq:wassersteinsum_new}
\sum_{(\hat{s}_0^i, \hat{g}^i) \in \hat{\mathcal{T}^*}} w((\hat{s}_0^i, \hat{g}^i), \tau^i) \text{.}
\end{equation}
Finally, from each of the $K$ selected trajectories $\tau^i$, the hindsight goal $g^i$ is selected from the state $s_t^i \in \tau^i$, that minimizes \eqref{eq:wasserstein_new}: 
\begin{equation}
\label{eq:hindsightgoal_new}
g^i = m\bigg (\argmin_{s_t^i \in \tau_i}\Big (d_G(\hat{g}^i - m(s_t^i)) - \frac{1}{L}V^\pi(s_0^i||m(s_t^i))\Big)\bigg).
\end{equation}
\textcolor{black}{The distance metric $d_G$} is based on shortest paths in a graph with a discrete representation of the goal space as vertices. 
The computation of shortest path distances is done pre-training and consists of creating a graph representing the environment and pre-computing shortest distances among vertices.

\textbf{{Graph Construction}}
\label{subsec:graph_construction}

\textcolor{black}{Let us consider an unbounded goal space $\mathcal{G}$ with an infinite number of goals;
let us further define a continuous but bounded subset of the goal space, the accessible goal space $\mathcal{G}_A \subset \mathcal{G}$. 
$\mathcal{G}_A$ contains all potential goals that the object can reach.
We then establish a representation of the accessible goal space $\mathcal{G}_A$ with an undirected graph.
A graph $G = (P, E)$ consists of a set of vertices $P$ with a set of weighted edges $E$, and an assigned weight $w$.
}
\begin{equation}
\label{eq:edgedef}
E \subset \{(p_1, p_2, w) \mid (p_1,p_2) \in P^2, p_1 \neq p_2 , w \in \mathbb{R}\},
\end{equation}
where $p_1$ and $p_2$ are two possible vertices.
In environments with obstacles, 
goals $g_{obs} \in \mathcal{G}$ lying within an obstacle that are blocked from being reached are not elements of the accessible goal space $g_{obs} \notin \mathcal{G}_A$.
%
Since $\mathcal{G}_A$ is bounded, it can be enclosed in a parallelepipedic bounding box defined by values $x_{min}$, $x_{max}$, $y_{min}$, $y_{max}$, $z_{min}$, $z_{max}$ $\in \mathbb{R}$, describing the span of this box in each coordinate direction. 
We then use this box to generate a finite set of vertices $\hat{P}$ spatially arranged as an orthorhombic lattice.
$\hat{P}$ is defined by the total number of vertices $n = n_x \cdot n_y \cdot n_z$, with $n_x, n_y, n_z \in \mathbb{N}$ in each coordinate direction of $\mathcal{G}_A$, or alternatively by the distance between two adjacent grid-points in each coordinate direction given by
%
\begin{equation}
\label{eq:deltas}
\Delta_x = \frac{x_{max}-x_{min}}{n_x-1}, \;
\Delta_y = \frac{y_{max}-y_{min}}{n_y-1}, \;
\Delta_z = \frac{z_{max}-z_{min}}{n_z-1}.
\end{equation}
Finally, we define the set of vertices as $P = \hat{P} \cap \mathcal{G}_A $,
where
\begin{equation}
\label{eq:vertices_2}
\begin{split}
\hat{P} := \{(x_{min} + \Delta_x \cdot i,\; y_{min} + \Delta_y \cdot j,\; z_{min} + \Delta_z \cdot k) \; \mid \; & i \in [0,n_x-1],  j \in [0, n_y-1], \\ & k \in [0, n_z-1]\}.
\end{split}
\end{equation}

\begin{figure}[htb!]
	\centering
	\begin{subfigure}[t]{.3\textwidth}
		\captionsetup{width=1\textwidth}
		\centering
		\includegraphics[width=0.8\linewidth, trim={3.1cm 1.5cm 3cm 3cm},clip]{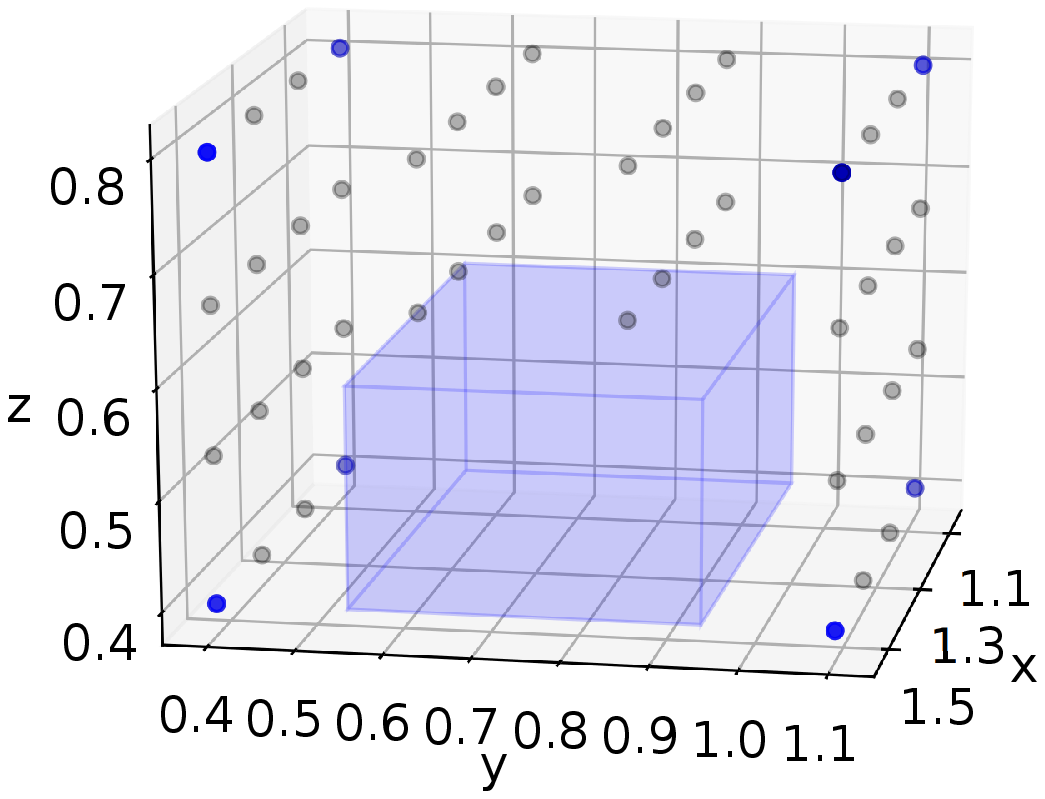}
		\caption{Creating vertices.}
		\label{fig:vertices}
	\end{subfigure}
	\begin{subfigure}[t]{.3\textwidth}
		\centering
		\captionsetup{width=1\textwidth}
		\includegraphics[width=0.8\linewidth, trim={3.3cm 1.5cm 3cm 3cm},clip]{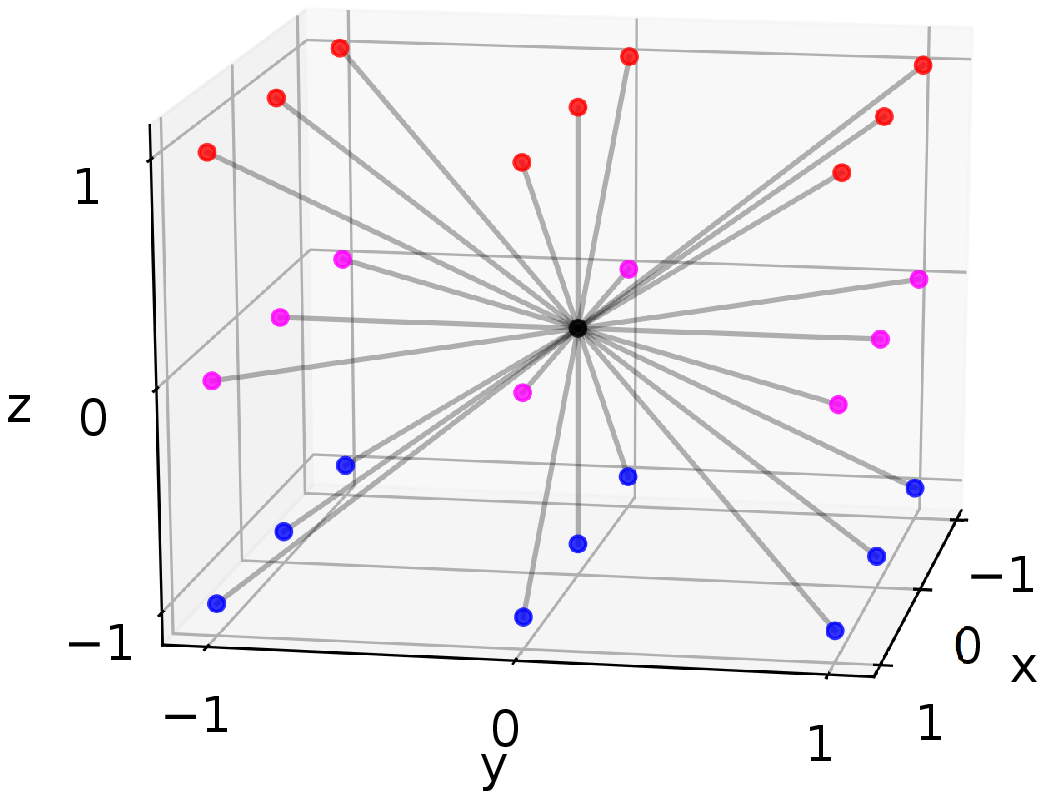}
		\caption{Creating outgoing edges.}
		\label{fig:connections}
	\end{subfigure}
	\begin{subfigure}[t]{.3\textwidth}
		\captionsetup{width=1.0\textwidth}
		\centering
		\includegraphics[width=0.8\linewidth, trim={3.1cm 1.5cm 3cm 3cm},clip]{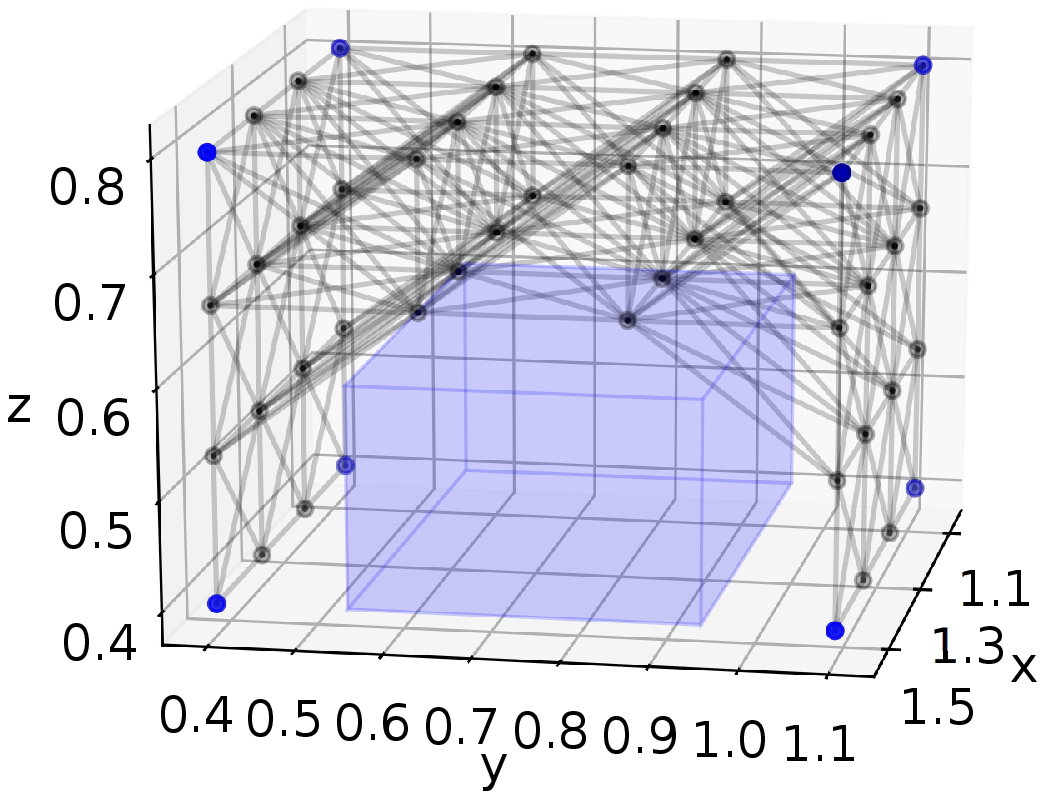}
		\caption{Creating graph.}
		\label{fig:graph2}
	\end{subfigure}
	\caption{Graph construction in the demo obstacle environment.}
	\label{fig:XXX}
\end{figure}

A set of vertices in a demo environment with an obstacle
is illustrated in Figure \ref{fig:vertices},
$\mathcal{G}_A$ is defined as the cuboid space marked by the blue balls, in which the obstacle is depicted by the blue box. $n_x = n_y = n_z = 4$. 
All the vertices are evenly distributed in $\mathcal{G}_A$ and no one lies inside the obstacle. 

As a next step, we connect two adjacent vertices $p_1 = (\hat{x}_1,\hat{y}_1,\hat{z}_1) \in P, \; p_2 = (\hat{x}_2, \hat{y}_2, \hat{z}_2) \in P$ with an edge of weight $w$ considering the following:
\begin{equation}
\label{eq:edges}
\begin{split}
(p_1, p_2, w) \in E \iff & |\hat{x}_2 - \hat{x}_1| \leq \Delta_x \; \text{ and }  |\hat{y}_2 - \hat{y}_1| \leq \Delta_y \; \text{ and }  |\hat{z}_2 - \hat{z}_1| \leq \Delta_z,
\end{split}
\end{equation}
with
$
\label{eq:weight}
w := \sqrt{(\hat{x}_2 -\hat{x}_1)^2+(\hat{y}_2-\hat{y}_1)^2+ \cdot (\hat{z}_2 - \hat{z}_1)^2}\text{.}
$

\begin{figure}[h!]
	\centering
	\begin{subfigure}[t]{.48\textwidth}
		\captionsetup{width=1.\textwidth}
		\centering
		\includegraphics[width=0.6\linewidth, trim={2.5cm 1.5cm 3cm 2.5cm},clip]{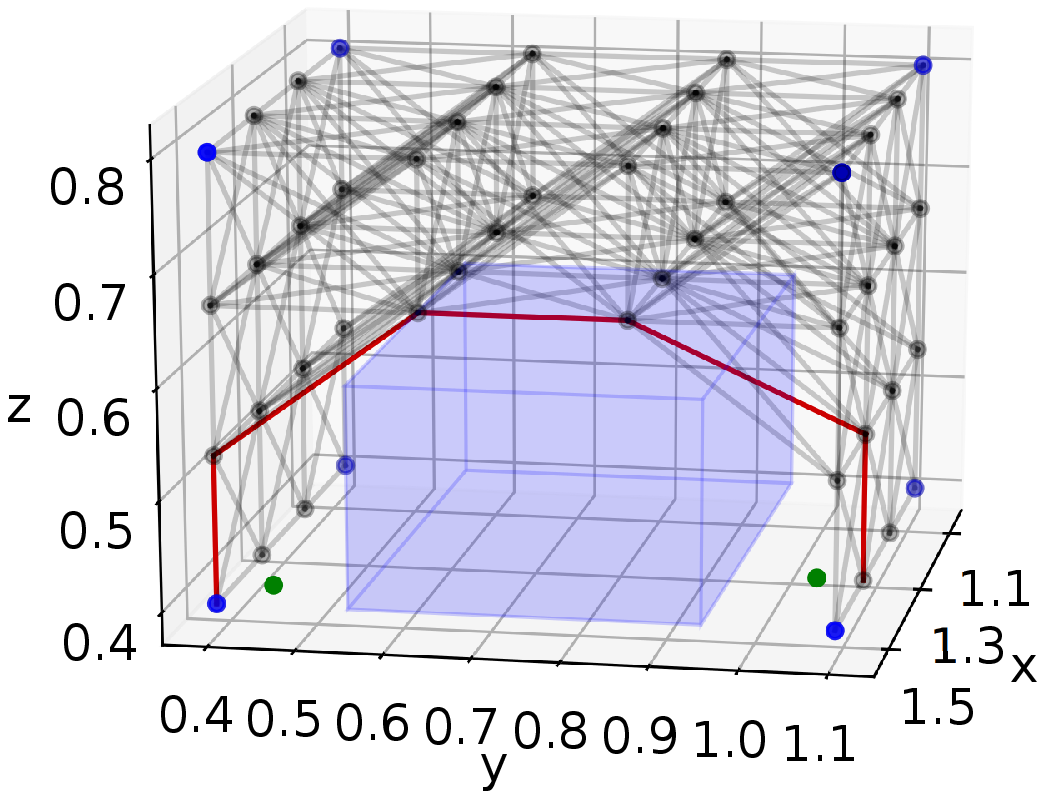}
		\caption{A: $d_g(g_1, g_2) = \hat{d}_G(\nu(g_1), \nu(g_2))=1.085$}
		\label{subfig:path_a}
	\end{subfigure}
	\begin{subfigure}[t]{.48\textwidth}
		\captionsetup{width=1.\textwidth}
		\centering
		\includegraphics[width=0.6\linewidth, trim={2.5cm 1.5cm 3cm 2.5cm},clip]{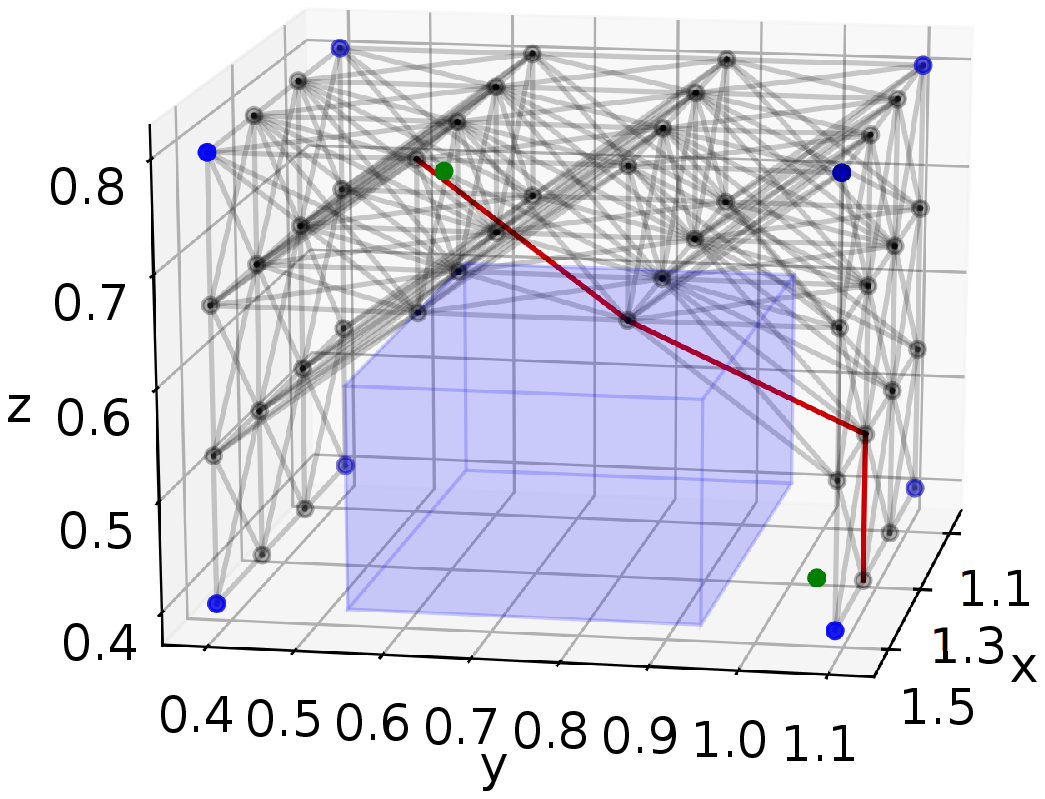}
		\caption{{B: $d_G(g_1, g_2) = \hat{d}_G(\nu(g_1), \nu(g_2))=0.718$}}
		\label{subfig:path_b}
	\end{subfigure}
	\caption{Graph-based distances and shortest paths (red lines) between two goals (green dots).}
	\label{fig:paths_1}
\end{figure}

In environments with obstacles, it is important to make sure that no edge in the graph cuts through an obstacle. 
\textcolor{black}{It should be noted that, for convex obstacles, we simply replace them with a minimal axis-aligned cuboid that can enclose the obstacle. 
We then remove this cuboid from $\mathcal{G}_A$; for a non-convex obstacle, we decompose it into several convex obstacles and then replace each of them with a minimal axis-aligned cuboid that can enclose the convex obstacle.}
Consequently, we require every convex sub-obstacle to be detected by at least one potential vertex $v \in \hat{P}$ but $v \notin \mathcal{G}_A$. 
Let us define the space of one of the cuboid with its edges $\alpha, \beta, \gamma$.
Thus, the graph must satisfy the graph density criterion \eqref{eq:vertexdensity} for every convex sub-obstacle, i.e., continuous set goals not included in $\mathcal{G_A}$:
\begin{equation}
\label{eq:vertexdensity}
\begin{split}
\Delta_x < \alpha_{min}^{obs};~
\Delta_y < \beta_{min}^{obs};~
\Delta_z < \gamma_{min}^{obs} 
\end{split},
\end{equation}
where $\alpha_{min}^{obs}$, $\beta_{min}^{obs}$, $\gamma_{min}^{obs}$ $\in \mathbb{R}$, describing the infimum length of edges of all the convex sub-obstacle.
%
When vertices are connected according to \eqref{eq:edges}, every vertex is connected to at most 26 adjacent vertices (as illustrated in Figure \ref{fig:connections}). 
However, an edge only exists as long as the adjacent vertex is contained in $\mathcal{G}_A$, thus, in case of an environment with obstacles, there is never an edge cutting through an obstacle as long as the graph density criterion \eqref{eq:vertexdensity} is satisfied. 
Figure \ref{fig:graph2} illustrates the final graph consisting of edges and vertices in our demo environment.

\begin{algorithm}[tb]
	\caption{Graph-Based Hindsight Goal Generation (G-HGG)}\label{alg:g-hgg}
	\begin{algorithmic}[1]
		\State \textbf{Given}: off-policy algorithm $\mathbb{A}$, sampling strategy $\mathbb{S}$, reward function $r_g: \mathcal{S} \times \mathcal{A} \rightarrow \mathbb{R}$
		\State Construct a graph $G=(V,E)$ as a discrete representation of $\mathcal{G}$ \Comment{section \ref{subsec:graph_construction}}
		\State Pre-compute shortest distances $\hat{d}_G$ between every pair of vertices $(p_1, p_2) \in P^2$ with Dijkstra 
		\State Initialize $\mathbb{A}$ and replay buffer $R$
		
		\For{$iteration$}
		\State Construct a set of $M$ intermediate tasks $\{(\hat{s}_0^i, g^i)\}_{i=1}^M$: \Comment{HGG}
		\begin{itemize}
			\item Sample target tasks $\{(\hat{s}_0^i, \hat{g}^i)\}_{i=1}^K \sim \mathcal{T}^*$ 
			\item Find $K$ distinct trajectories $\{\tau^i\}_{i=1}^K$ that together minimize \eqref{eq:wassersteinsum_new} \Comment{weighted bipartite matching, based on $d_G \approx \hat{d}_G$}
			\item Find $M$ intermediate tasks $(\hat{s}_0^i, g^i)$ by selecting intermediate goal $g^i$ from each $\tau^i$ according to \eqref{eq:hindsightgoal_new} \Comment{based on $d_G \approx \hat{d}_G$}
		\end{itemize}
		\For{$episode = 1,M$}
		\State $(s_0,g) \leftarrow (\hat{s}_0^i, g^i)$ \Comment{hindsight goal-oriented exploration}
		\For{$t = 0, T-1$}
		\State Sample an action $a_t$ using the policy from $\mathbb{A}$ with noise:
		$
		a_t \leftarrow \pi(s_t||g) + \mathcal{N}_t 
		$
		\State Execute the action $a_t$ and observe a new state $s_{t+1}$
		\EndFor
		
		\For{$t = 0, T-1$}
		\State $r_t := r_g(s_t,a_t)$;~Store transition $(s_t||g, a_t, r_t, s_{t+1}||g)$ in $R$; \Comment{experience replay}
		\State Sample a set of additional goals for replay $G:=\mathbb{S}(current\, episode)$
		\For{$g' \in G$}
		\State $r':=r_{g'}(s_t, a_t)$;~ Store the transition $(s_t||g', a_t, r', s_{t+1}||g')$ in $R$; \Comment{HER}
		\EndFor
		\EndFor
		\EndFor
		\For{$t = 1, N$}
		\State Sample a minibatch $B$ from the replay buffer $R$ \Comment{HER or EBP}
		\State Perform one step of optimization using $\mathbb{A}$ and minibatch $B$ using DDPG
		\EndFor
		\EndFor
		
	\end{algorithmic}
\end{algorithm}

\textbf{Shorted Distance Computation}
Considering the created graph that represents the environment, we can employ a shortest path algorithm such as Dijkstra's algorithm \cite{Dijkstra1959} to calculate shortest paths and shortest distances $\hat{d}_G$ between every possible pair of vertices $(p_1, p_2) = ( (\hat{x}_1, \hat{y}_1, \hat{z}_1), (\hat{x}_2, \hat{y}_2, \hat{z}_2) ) \in P^2$ in a graph $G = (P,E)$. 
All possible combinations of the resulting shortest distance function $\hat{d}_G$ can be efficiently pre-computed with Dijkstra and stored in an $n \times n$ table, where $n$ denotes the number of vertices in $P$.
An ablation study of $n$ can be found in the Appendix~\ref{subsubsec_vertices}.

Given two goals $g_1 = (x_1, y_1, z_1) \in \mathcal{G}$, $g_2 = (x_2, y_2, z_2) \in \mathcal{G}$ and a graph $G = (P,E)$ with representing the approximate goal space $\mathcal{G}_A \subset \mathcal{G}$ with $x_{min}$, $x_{max}$, $y_{min}$, $y_{max}$, $z_{min}$, $z_{max}$, $\Delta_x$, $\Delta_y$, and $\Delta_z$, the graph-based distance $d: \mathcal{G}^2 \rightarrow \mathbb{R}$ is defined such that
\begin{equation}
\label{eq:distance_G_A}
d_G(g_1, g_2) = 
\begin{cases} 
\hat{d}_G\Big(\nu(g_1), \nu(g_2)\Big), & \text{if } g_1 \in \mathcal{G}_A \land g_2 \in \mathcal{G}_A \\ 
\infty, & \text{otherwise}
\end{cases}
\end{equation}
where $\nu: \mathcal{G}_A \rightarrow P$ maps goals in $\mathcal{G}_A$ to the closest vertex in $P$:
\begin{equation}
\label{eq:nu}
\begin{split}
& \nu(g) = \nu(x,y,z) = (\hat{x}, \hat{y}, \hat{z}) = \\
& \big(x_{min} + \Delta_x \cdot \Big\lfloor \frac{x-x_{min}}{\Delta_x} \Big\rceil,~ 
y_{min} + \Delta_y \cdot \Big\lfloor \frac{y-y_{min}}{\Delta_y} \Big\rceil,~
z_{min} + \Delta_x \cdot \Big\lfloor \frac{z-z_{min}}{\Delta_z} \Big\rceil 
\big)
\end{split}
\end{equation}
$\lfloor a \rceil$ rounds any $a \in \mathbb{R}$ to the closest integer value.
Figure \ref{fig:paths_1} illustrates a visualization of graph-based goal distance computation between goals $g_1 \in \mathcal{G}_A$, $g_2 \in \mathcal{G}_A$. 
In both Cases A (Figure \ref{subfig:path_a}) and B (Figure \ref{subfig:path_b}), both goals are contained in the accessible goal space $\mathcal{G}_A$. Therefore, the graph-based distance is approximated by the shortest distance between the two vertices $\nu(g_1)$ and $\nu(g_2)$ that are closest to the goals $g_1$ and $g_2$, respectively. 
Mathematically, $d(g_1,g_2) = \hat{d}_G(\nu(g_1), \nu(g_2))$.


\textbf{Algorithm and Stop Condition:}
The overall G-HGG algorithm is provided as Algorithm \ref{alg:g-hgg}. 
The main differences between HGG and G-HGG are in \textcolor{black}{steps 2,3 and 6}. 
Step 2 (graph construction) and step 3 (shortest distance computation) are performed pre-training. 
The pre-computed table of shortest distances $\hat{d}_G$ is used in step 6.
%
%
%
G-HGG guides exploration by hindsight goals. 
This is especially useful at the beginning of training, where hindsight goals are far from the target goals $ \mathcal{G}_T$. 
When training progresses and a certain fraction $\delta_{stop} \in [0,1]$ of the hindsight goal candidates are very close to sampled target goals, stopping HGG and continuing with vanilla HER will lead to faster learning and higher success rates.
Because of the page limit, we include the ablation study on $\delta_{stop}$ in Appendix~\ref{chapter:ablation}.


%
%
%
\section{Experiments}
\label{chapter:experiments}
%

To demonstrate the advantages of G-HGG over HGG and vanilla HER, we create new experimental environments based on the standard robotic manipulation environments from OpenAI Gym \cite{Plappert2017}.
%
	
\begin{figure}[htb!]
	\centering
	\captionsetup{skip=1pt}
	\begin{subfigure}[t]{.24\textwidth}
		\centering
		\includegraphics[width=0.75\linewidth]{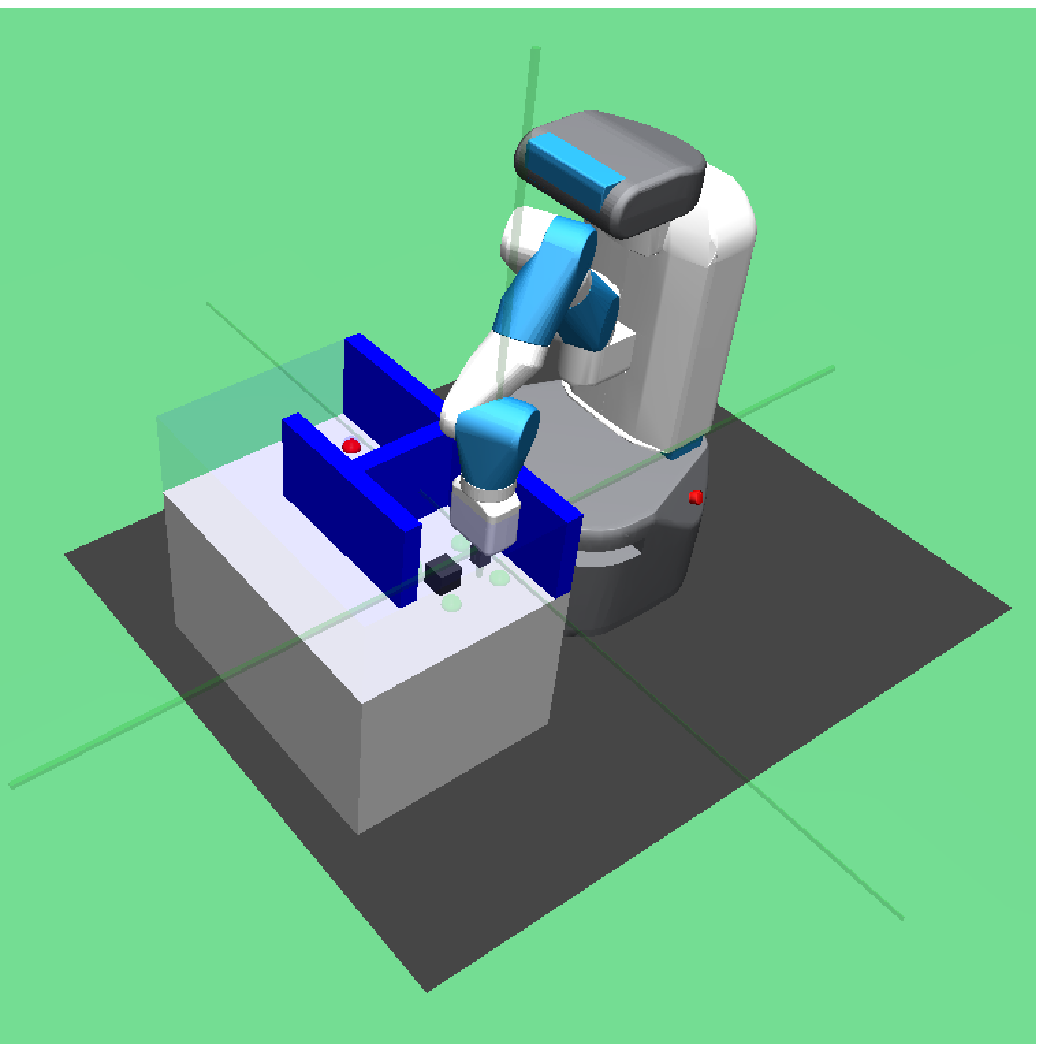}
		\caption{FetchPushLabyrinth}
		\label{fig:labyrinth}
	\end{subfigure}
	\begin{subfigure}[t]{.24\textwidth}
		\centering
		\includegraphics[width=0.75\linewidth]{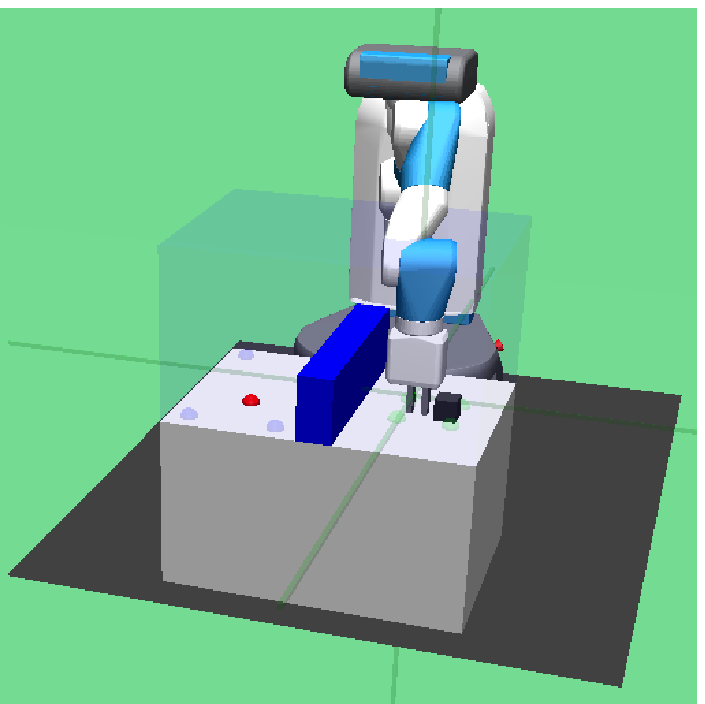}
		\caption{FetchPickObstacle}
		\label{fig:obstacle}
	\end{subfigure}
	\begin{subfigure}[t]{.24\textwidth}
		\centering
		\includegraphics[width=0.75\linewidth]{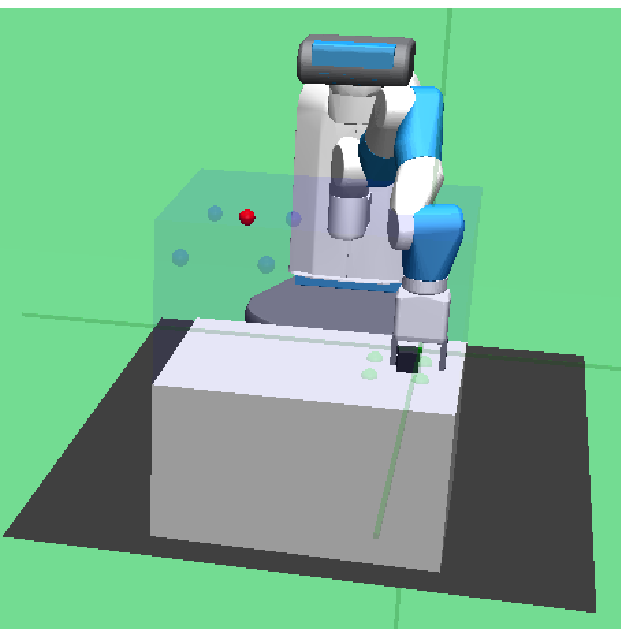}
		\caption{FetchPickNoObstacle }
		\label{fig:noobstacle}
	\end{subfigure}
	\begin{subfigure}[t]{.24\textwidth}
		\centering
		\includegraphics[width=0.75\linewidth]{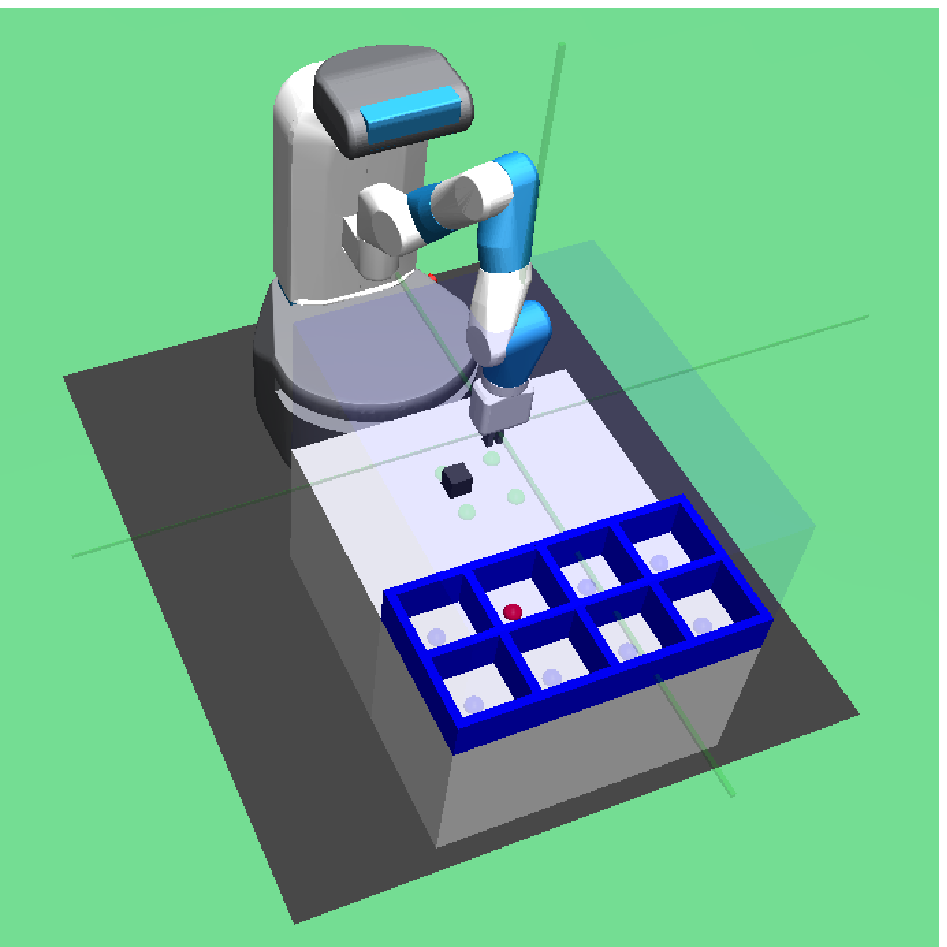}
		\caption{FetchPickAndThrow}
		\label{fig:throw}
	\end{subfigure}
	\caption{Robotic manipulation environments.}
	\vspace{-1.0em}
	\label{fig:envs}
\end{figure}

\begin{itemize}[leftmargin=*, noitemsep, topsep=0pt]
	\itemsep0em
	\item \textbf{FetchPushLabyrinth} (Figure \ref{fig:labyrinth}): the goal is to push the cube from its initial position around the blue obstacles (labyrinth) to a target goal. The gripper remains permanently closed. 
	
	\item \textbf{FetchPickObstacle} (Figure \ref{fig:obstacle}): the goal is to pick up the cube from its initial position, lift it over the obstacle, and place it at a target goal position. Gripper control is enabled.
	
	\item \textbf{FetchPickNoObstacle} (Figure \ref{fig:noobstacle}): the goal is to pick up the cube from its initial position, lift it up, and place it at a target goal position. No obstacle is present in this scenario, but the target goals are located in the air. Gripper control is enabled. 
	
	\item \textbf{FetchPickAndThrow} (Figure \ref{fig:throw}): the goal is to pick up the cube from its initial position, lift it up, and throw it into one of the eight boxes (obstacles). Gripper control is enabled.
	
\end{itemize}
%
%
%
%
%
%
We tested G-HGG on the four environments to compare its performance to HGG and HER. 
Since we know from \cite{Zhao2018} and \cite{Ren2019} that energy-based prioritization (EBP) \cite{Zhao2018} significantly enhances the performance of both HER and HGG, we used EBP in all our trainings of HER, HGG, and G-HGG. 
The results clearly show that G-HGG outperforms HGG by far in environments with obstacles, both in terms of sample efficiency and maximum success rate. 
In the environment without obstacles, the performance of G-HGG was still comparable to HGG in terms of sample efficiency and maximum success rate.

\begin{figure}[htb!]
	\centering
	\captionsetup{skip=0pt}
	\begin{subfigure}[t]{.255\textwidth}
		\captionsetup{skip=0pt}
		\centering
		\includegraphics[width=1\linewidth, trim={0cm 0.0cm 0.2cm 0.1cm},clip]{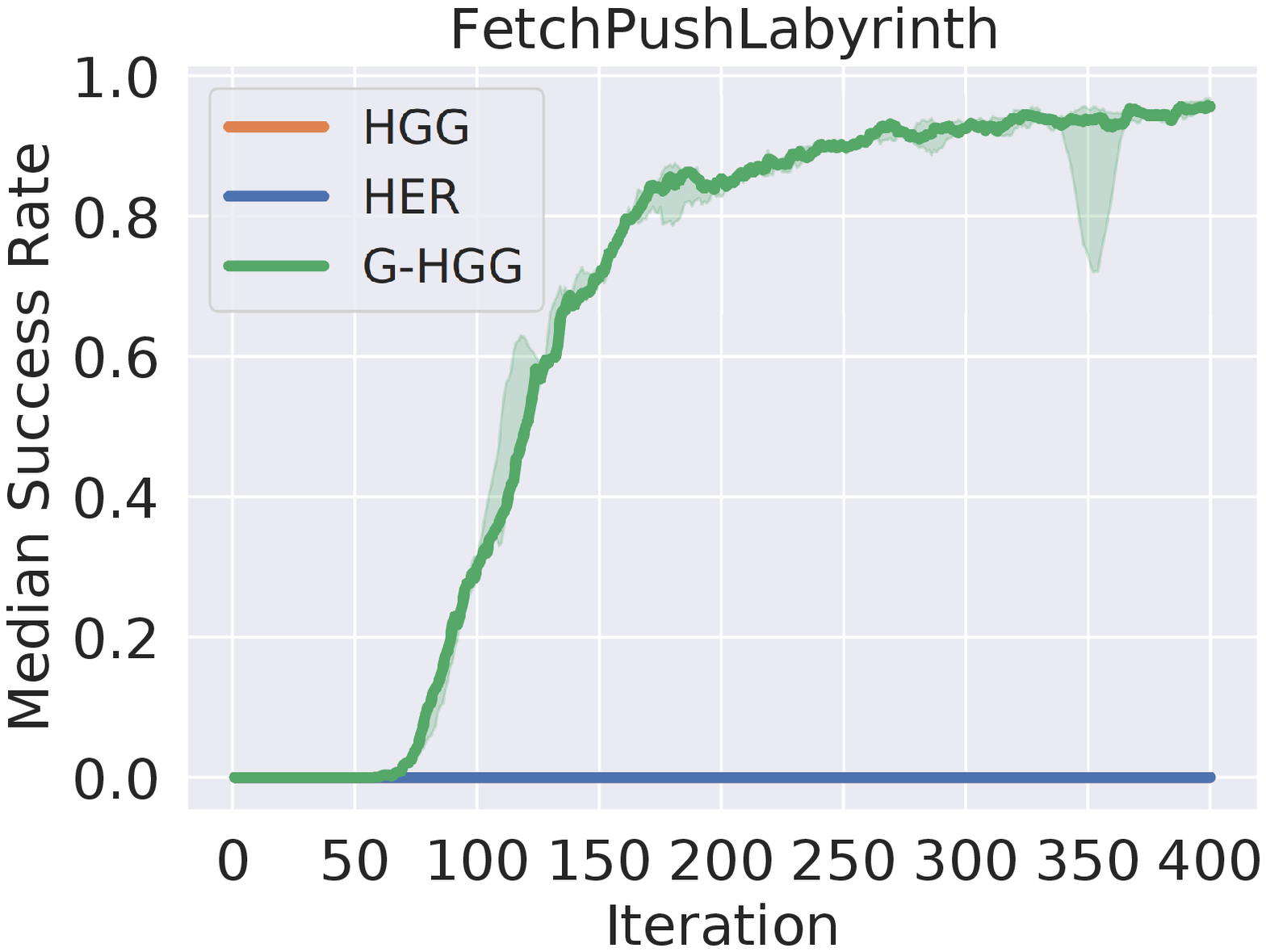}
		\label{subfig:results_labyrinth}
	\end{subfigure}
	\begin{subfigure}[t]{.24\textwidth}
		\captionsetup{skip=0pt}
		\centering
		\includegraphics[width=1\linewidth, trim={1.0cm 0.0cm 0.2cm 0.0cm},clip]{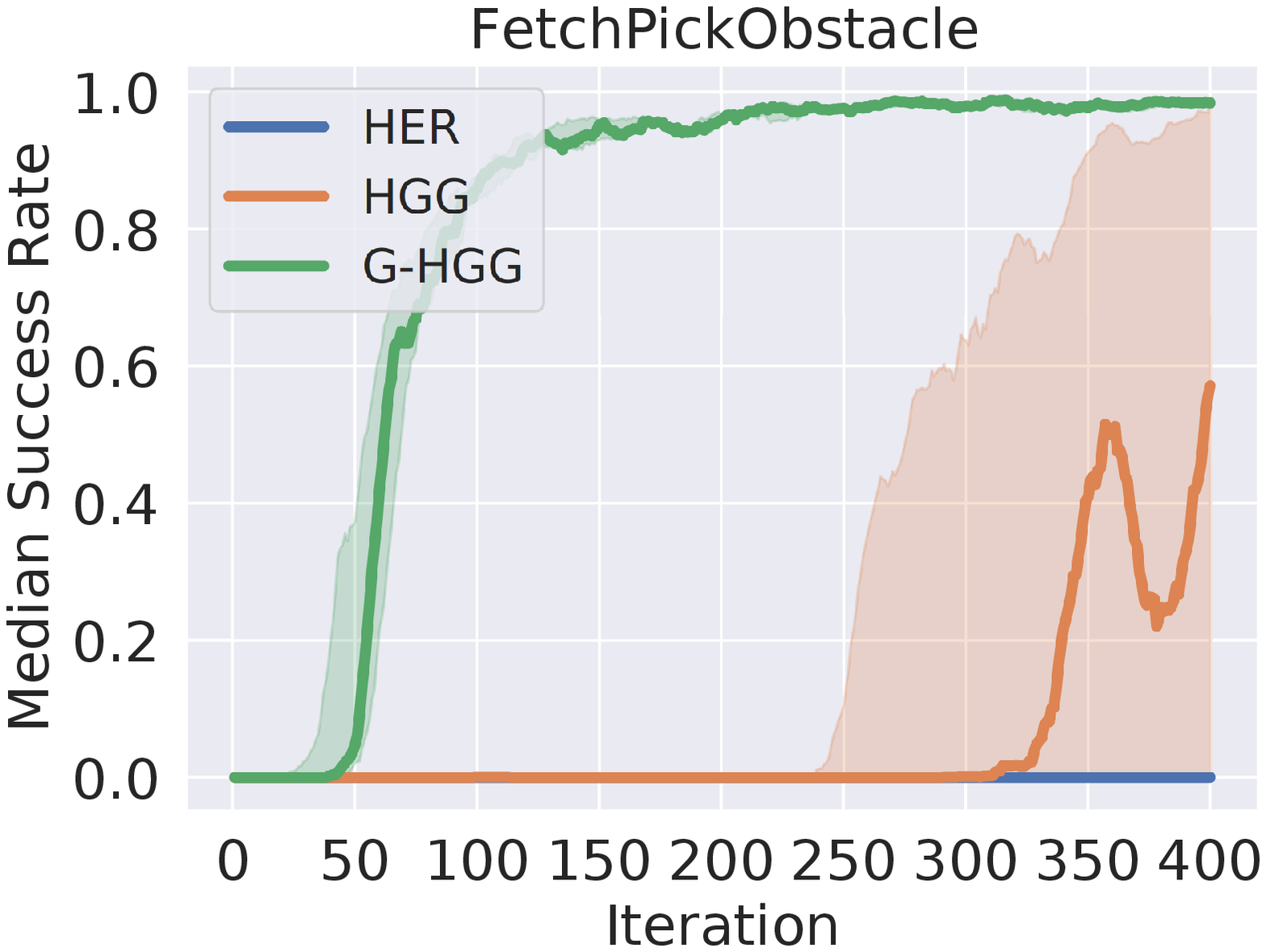}
		\label{subfig:results_obstacle}
	\end{subfigure}
	\begin{subfigure}[t]{.24\textwidth}
		\captionsetup{skip=0pt}
		\centering
		\includegraphics[width=1\linewidth, trim={1cm 0.0cm 0.2cm 0.0cm},clip]{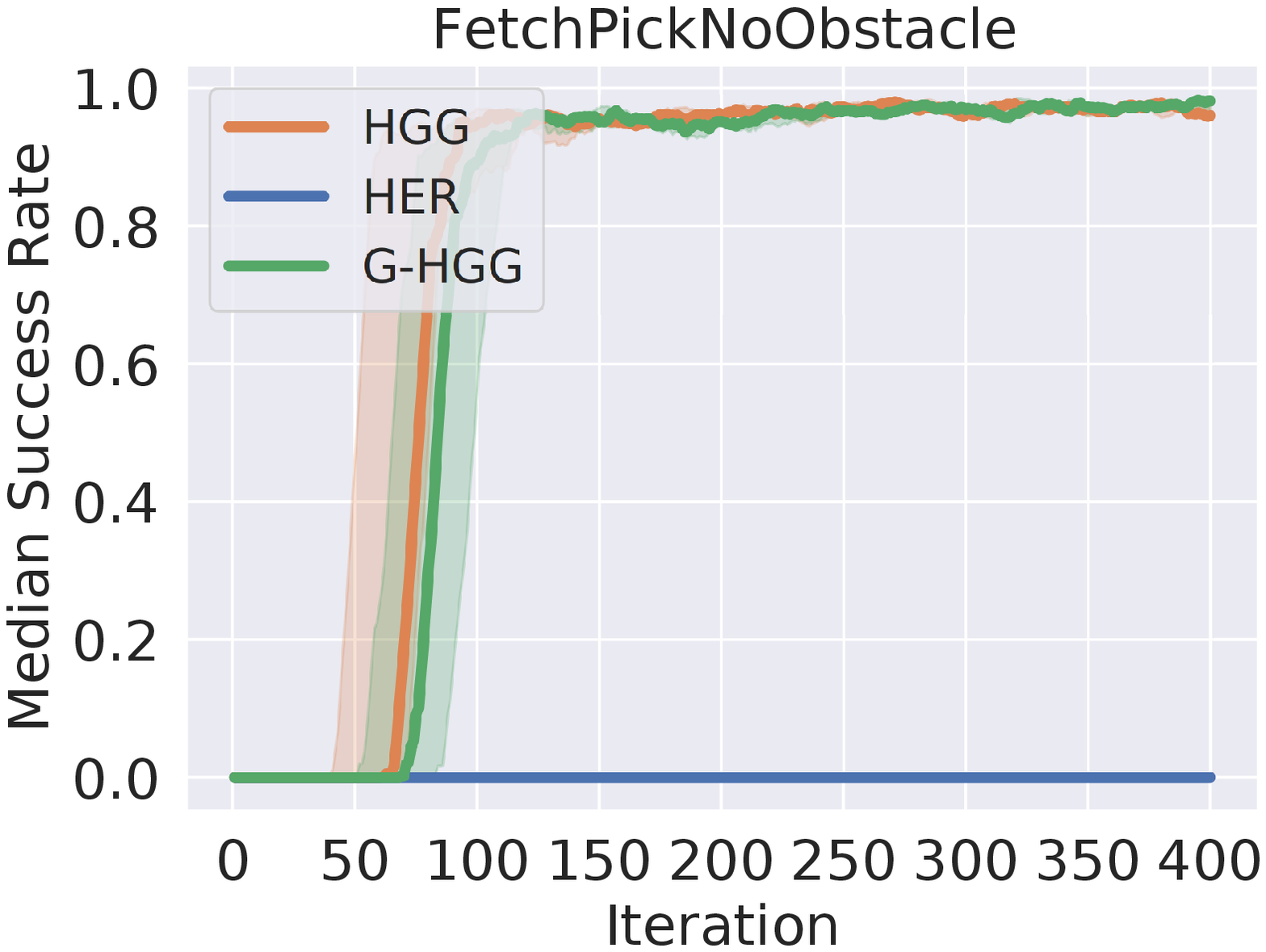}
		\label{subfig:results_noobstacle}
	\end{subfigure}
	\begin{subfigure}[t]{.24\textwidth}
		\captionsetup{skip=0pt}
		\centering
		\includegraphics[width=1\linewidth, trim={1cm 0.0cm 0.2cm 0.0cm},clip]{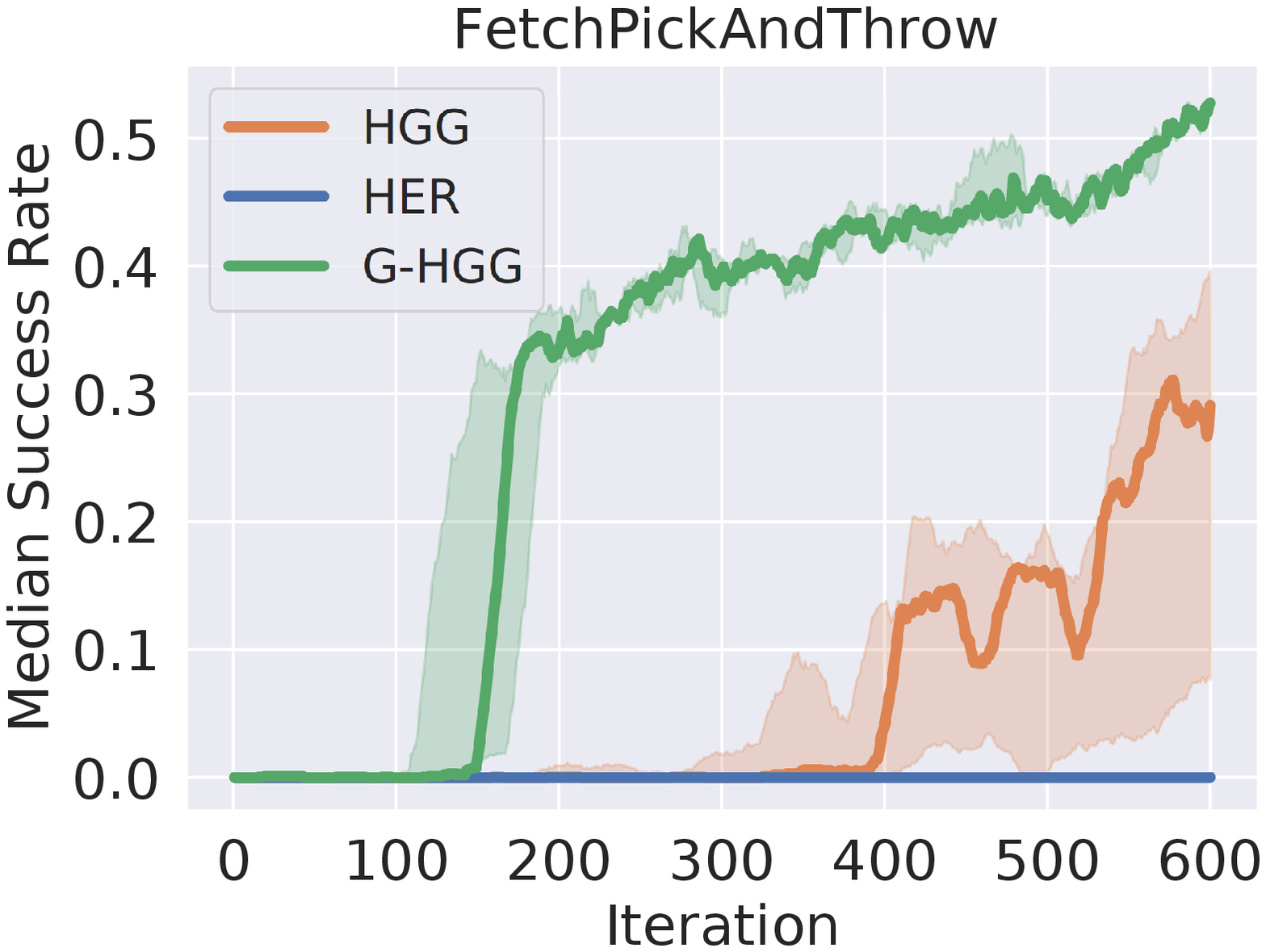}
		\label{subfig:results_throw	}
	\end{subfigure}
	\caption{Median test success rate (line) and interquartile range (shaded) of G-HGG, HGG, and HER.}
	\vspace{-0.5em}
	\label{fig:results}
\end{figure}
\begin{figure}[htb!]
	\centering
	\begin{subfigure}[t]{.48\textwidth}
		\centering
		\includegraphics[width=0.55\linewidth, trim={3cm 1cm 1cm 2cm},clip]{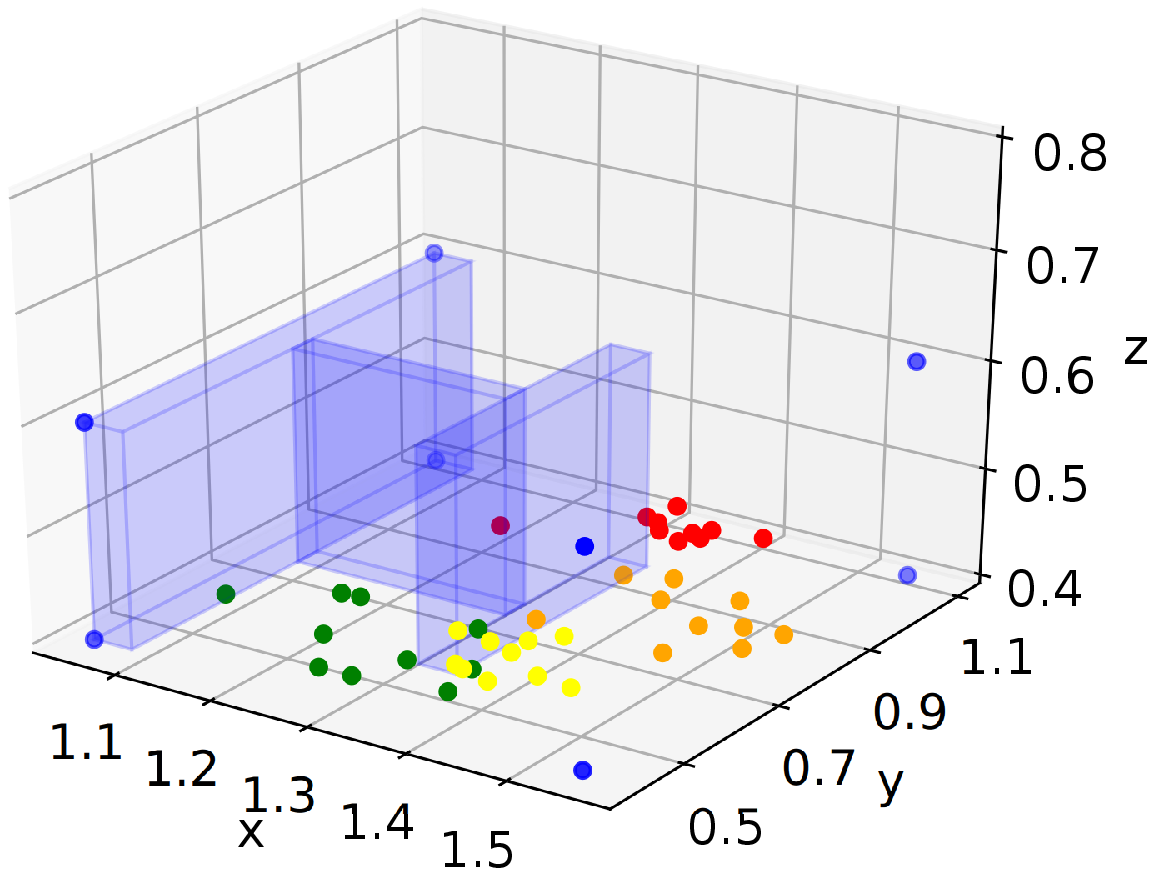}
		\label{subfig:goals_labyrinth_ghgg}
	\end{subfigure}
	\begin{subfigure}[t]{.48\textwidth}
		\centering
		\includegraphics[width=0.55\linewidth, trim={3cm 1cm 1cm 2cm},clip]{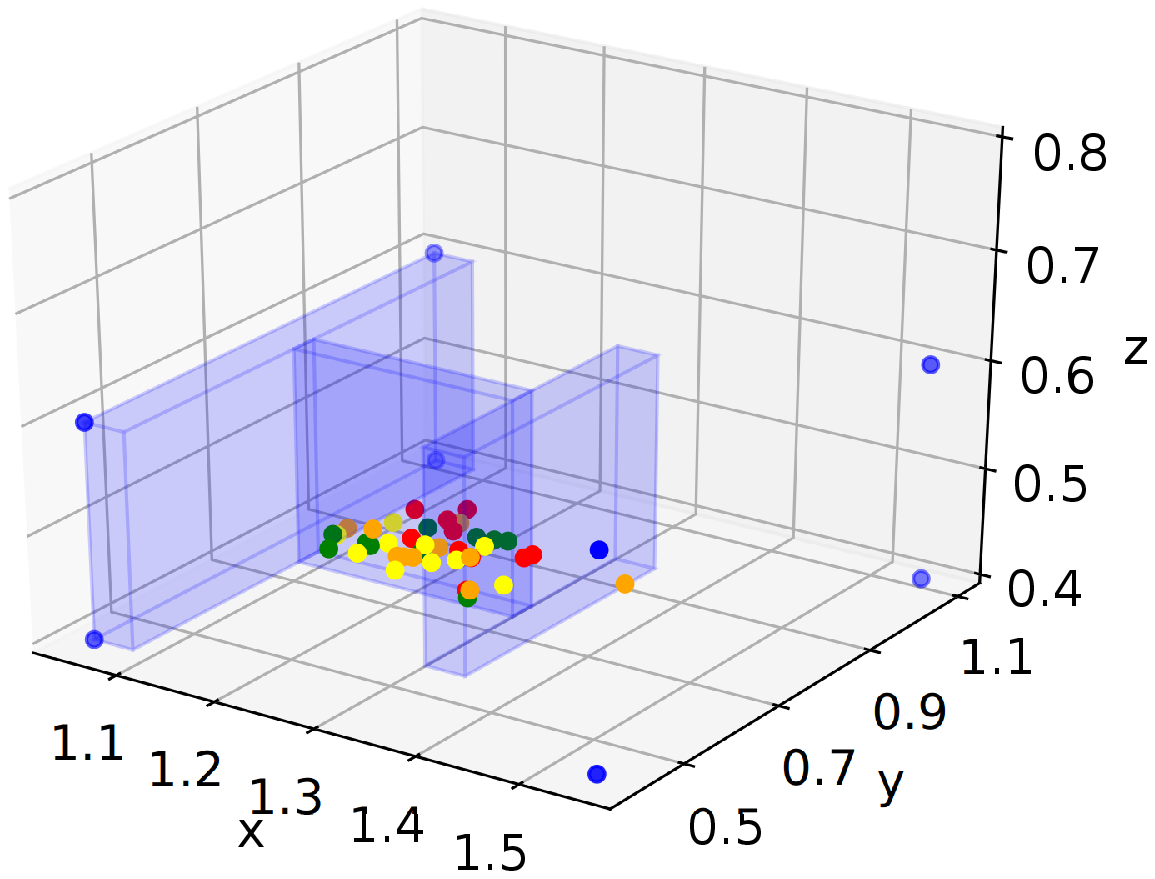}
		\label{subfig:goals_labyrinth_hgg}
	\end{subfigure}
	\caption{Hindsight goals in FetchPushLabyrinth after 20 (red), 40 (orange), 60 (yellow) and 80 (green) episodes of G-HGG (Left) / HGG (Right). One iteration contains $M = 50$ episodes.}
	\label{fig:goals_labyrinth}
	\vspace{-0.5em}
\end{figure}

Figure \ref{fig:results} shows success rates of G-HGG, HGG, and HER (median and interquartile range of five training runs each), plotted over training iterations in four environments. 
The most remarkable results can be observed in the FetchPushLabyrinth environment. 
While HER and HGG display no success over 400 iterations, G-HGG reaches a success rate of $80\%$ after 170 iterations, increasing to over $90\%$ after 300 iterations. 
By comparing a sample of hindsight goals from iterations 20, 40, 60, and 80 (Figure \ref{fig:goals_labyrinth}), it becomes obvious that G-HGG outperforms HGG by far. 
While the graph-based distance metric used in G-HGG leads to a choice of hindsight goals guiding the agent around the obstacle towards the target goals , HGG repeatedly uses goals that are closest to the target goals with respect to the euclidean metric. 
Since this euclidean metric based shortest path is blocked by an obstacle, the agent gets stuck trying to reach the hindsight goals (pushing the object against the obstacle), never achieving more promising goals, and thus, never reaching the target goals.

In the FetchPickObstacle environment, G-HGG outperforms HGG in terms of sample efficiency, due to graph-based distances supporting goals that avoid the obstacle. 
Since the euclidean distance metric is at least partly (in x and y direction) valid, HGG eventually achieves a notable success rate as well. 
However, the plots show that there is a large variance within the HGG trainings, emphasizing the disadvantage of random exploration in HGG with non-meaningful hindsight goals over guided exploration in G-HGG. 
Vanilla HER cannot solve the task.

FetchPickNoObstacle is an environment where G-HGG has no advantage over HGG. 
As no obstacle is present, the euclidean metric is applicable to compare distances, allowing HGG to perform well. 
Since the euclidean metric used in HGG is more exact than the graph-based distances in G-HGG, it is not surprising that HGG yields slightly better training results. 
However, the similarity of the curves shows that G-HGG's performance is comparable to HGG in terms of sample efficiency and success rate even in environments without obstacles. 
Overall, we assume that G-HGG is generally applicable to all environments where HGG yields good training results. 
HER still fails to reach the target goals. 

FetchPickAndThrow is a difficult task, since target goals are not uniformly sampled from a continuous target goal distribution, but from a discrete set of eight goals. 
G-HGG yields better results than HGG in terms of success rate after 600 iterations and is clearly more sample efficient. 
Both HGG and G-HGG cannot achieve success rates above $60\%$ due to the difficulty of the task, involving picking the object, lifting it up, and dropping it while giving it a well-dosed push in the desired direction. 
Since the final policy trained to achieve the task results from guided, but still random exploration, it is no surprise that the agent cannot develop a perfect throwing motion from sparse rewards. 

\section{Related Work}

Informative and effective explorations are essential for solving goal-conditioned RL tasks with sparse rewards. 
An amount of research has emerged and we briefly discuss them from three main ideas.

\textbf{Prioritized Experience Replay}
One major drawback of HER has been its inefficient random replay of experience. 
Research has shown that prioritized sampling of transitions from the replay buffer significantly increases sample efficiency. 
Prioritized sampling can be based on the TD-error \cite{Schaul2016}, reward-weighted entropy \cite{Zhao2019}, transition energy \cite{Zhao2018}, and density of achieved goals \cite{Zhao2019a}. \\
%
%
\textbf{Curriculum Learning}
Another way to tackle the exploration problem in sparse-reward multi-goal RL is curriculum learning (CL)~\cite{Riedmiller2018}, which presents problems in a favorable order, the so-called curriculum. 
%
%
One CL approach to improve exploration is augmenting the sparse learning problem with basic, easy-to-learn auxiliary tasks \cite{Riedmiller2018, Eppe2019}. 
In the absence of extrinsic motivation due to the sparsity of external rewards, intrinsic motivation can be used to create a curriculum for improved exploration \cite{Singh2005, Pathak2017, Forestier2017, Pere2018, Sukhbaatar2018, Colas2019, Aubret2019}. 
Another way of constructing a meaningful curriculum is to predict high-reward states and generate goals close to these meaningful states \cite{Goyal2019,Florensa2018,Florensa2017, Fang2019}. 
%
\\
\textbf{Representation Learning}~\cite{6472238} is a promising framework to solve goal-conditioned RL tasks, in which representative abstractions are interpreted from high-dimensional observations to low-dimensional latent states.  
With the representations as a planner, model-free RL algorithms are able to perform control tasks. 
Some work showed that learned representations can be used to solve navigation and goal-reaching tasks for mobile agents. 
Robotic manipulations tasks were implemented but limited to 2D space because most representations were learned from images~\cite{srinivas2018universal, Ghosh2019LearningAR,eysenbach2019search}.  

\section{Conclusion}

We proposed a novel automatic hindsight goal generation algorithm G-HGG on the basis of the HGG for robotic object manipulation in environments with obstacles, by which the selection of valuable hindsight goals is generated with a graph-based distance metric. 
We formulated our solution as a graph construction and shortest distance computation process as pre-training steps. 
Experiments on four different challenging object manipulation tasks demonstrated superior performance of G-HGG over HGG and HER in terms of both maximum success rate and sample efficiency. 
Future research could concentrate on improvement, extension, and real-world deployment of G-HGG.

\section*{Broader Impact}

\bibliographystyle{plain}
\bibliography{bing_neurips_2020}

\end{document}